\documentclass[10pt,twocolumn,letterpaper]{article}

\usepackage{cvpr}
\usepackage{times}
\usepackage{epsfig}
\usepackage{graphicx}
\usepackage{amsmath}
\usepackage{amssymb}

\usepackage[breaklinks=true,bookmarks=false]{hyperref}

\cvprfinalcopy 

\def\cvprPaperID{****} 

\begin{document}

\title{Road Mapping in Low Data Environments with OpenStreetMap}

\author{John Kamalu\thanks{Equal contribution.}\\
Department of Computer Science\\
Stanford University\\
{\tt\small jkamalu@cs.stanford.edu}
\and
Benjamin Choi\footnotemark[1]\\
Department of Electrical Engineering\\
Stanford University\\
{\tt\small benchoi@stanford.edu}
}

\maketitle

\begin{abstract}
    Roads are among the most essential components of any country's infrastructure. By facilitating the movement and exchange of people, ideas, and goods, they support economic and cultural activity both within and across local and international borders. A comprehensive, up-to-date mapping of the geographical distribution of roads and their quality thus has the potential to act as an indicator for broader economic development. Such an indicator has a variety of high-impact applications, particularly in the planning of rural development projects where up-to-date infrastructure information is not available. This work investigates the viability of high resolution satellite imagery and crowd-sourced resources like OpenStreetMap \cite{OpenStreetMap} in the construction of such a mapping. We experiment with state-of-the-art deep learning methods to explore the utility of OpenStreetMap data in road classification and segmentation tasks. We also compare the performance of models in different mask occlusion scenarios as well as out-of-country domains. Our comparison raises important pitfalls to consider in image-based infrastructure classification tasks, and shows the need for local training data specific to regions of interest for reliable performance.

\end{abstract}

\section{Introduction}
    OpenStreetMap (OSM) is a crowd-sourced map with street-level labels made available to the public for unrestricted usage and contribution. OpenStreetMap relies on the wealth of local expertise which it crowd-sources to form a patchwork of local and regional infrastructure information. New contributors are presumed to possess an expertise which supersedes that of their predecessors such that they are able to add, modify, or delete existing data that they deem accurate. This assumption strikes the balance between coverage and precision \cite{mooney2017openstreetmap}. In exchange for an inexpensive, complete map of the world's road infrastructure, we might expect the crowd-source contribution mechanism to introduce some added variance to the data; the distribution of classification labels may vary across contributors, and the precision of localization labels (GPS) may vary with the instrument used to perform the positioning. However, the accessibility of OSM data in areas with poor data integrity renders invaluable the prospect of an algorithm capable of rapid and accurate mapping of such regions at little to no cost.
    
    This work is undertaken in association with \textbf{Natural Capital Project} and \textbf{The Nature Conservancy}, and serves to investigate the viability of OSM data in the development of a useful algorithm. The long-term goal of this work is defined as the inexpensive identification and classification of roads in low data environments over time. Specifically, we aim to create a historical map capable of tracking the construction, development, and deterioration of transportation infrastructure, especially where this kind of information is lacking and could be of great service to policy makers and conservationists alike. In this report, we seek to find the limitations of OSM data from two countries, Kenya and Peru, via classification and segmentation experiments with high-performing deep convolutional models. We report evaluation metrics and present analyses of our results that provide a path forward for continued research in infrastructure classification.

\section{Related Work}

This work relies on several previous explorations of related domains, taking great inspiration from similar deep learning techniques applied to other infrastructure-related tasks.

\subsection{Deep Learning for Infrastructure Classification}

Modern deep learning methods have been applied to a variety of satellite imagery based classification tasks for the purpose of mapping global infrastructure development. Deep learning has been used to predict binary infrastructure quality metrics (electricity, sewage, piped water, road quality, etc., from Afrobarometer 6 survey data) using Landsat 8/Sentinel 1 satellite imagery data and a wide range of auxiliary datasets (nightlights, Open Street Maps) \cite{afrobarometer}. Models trained to predict infrastructure quality based on satellite imagery perform considerably better than those trained on OSM or nightlights based on AUROC metrics. Surprisingly, models trained on Open Street Maps data are shown to perform worse than models trained on Landsat 8 data on road quality classification tasks. 

In contrast to employing a static model on a wide variety of datasets as in \cite{afrobarometer}, \cite{Cadamuro:2019:SSM:3314344.3332493} employs a wide range of machine learning techniques ranging from CNNs to autoencoders and spatial LSTMs to perform 5-class road quality classification tasks. Using extensive datasets from Kenya based on a precise road quality index (IRI) and DigitalGlobe WorldView satellite images. Autoencoders perform particularly well in generalization; the unsupervised approach allows for full use of data in order to generate meaningful features, while other methods are only able to take advantage of a subset of the full dataset due to the elimination of WorldView/IRI data “(x, y) pairs” that were too temporally distant to reliably be deemed as more or less the “same road.” 

\subsection{Road Segmentation}

Recent advances in deep learning have greatly aided performance on road detection tasks consisting of the labeling of roads in high resolution satellite imagery. For example, \cite{DBLP:journals/corr/abs-1711-10684} achieves high performance on semantic segmentation of roads in satellite imagery using a novel architecture known as a Residual U-Net, a variant on an architecture normally used for biomedical segmentation tasks \cite{DBLP:journals/corr/RonnebergerFB15}. The U-Net features residual units that aim to propagate signal from low, pixel-level details upward to higher level features, a strategy that allows for highly-detailed segmentation. Using a set of satellite images of Massachusetts roads, the extraction method achieves higher precision and recall metrics than other modern road detection methods. Qualitatively, road detection does very well, achieving low-noise and clean roads even in cases road segments are partially-occluded. Closely related to our task is \cite{Nachmany_2019_CVPR_Workshops}, which aims to detect roads in the developing world based on satellite imagery. 

Road extraction has received significant attention in recent years and has featured in several major competitions, e.g. the 2018 DeepGlobe Road Extraction Challenge hosted by CodaLab (DeepGlobe 2018). This challenge uses 50 cm resolution satellite imagery from DigitalGlobe at 1024 x 1024 pixels for a total of some 6K labeled images; road segmentation is evaluated with respect to IoU (pointwise intersection over union). The first place solution uses the D-LinkNet architecture, a deep convolutional encoder/decoder architecture which modifies the original LinkNet model to leverage dilated convolution to perform semantic segmentation. The encoder -- a pretrained ResNet34 architecture -- and the decoder -- a series of upsampling convolutional operations with depth-distributed residual connections to the encoder -- are linked via dilated convolution to enlarge the receptive field of feature points \cite{Zhou2018DLinkNetLW, abhishek2017linknet, yu2015multi}. The authors of this solution publish their code in addition to trained model weights, which we make use when evaluating the quality of segmentation masks extracted from our data.

\section{Data}

\begin{figure}
\centering
\includegraphics[width=\linewidth]{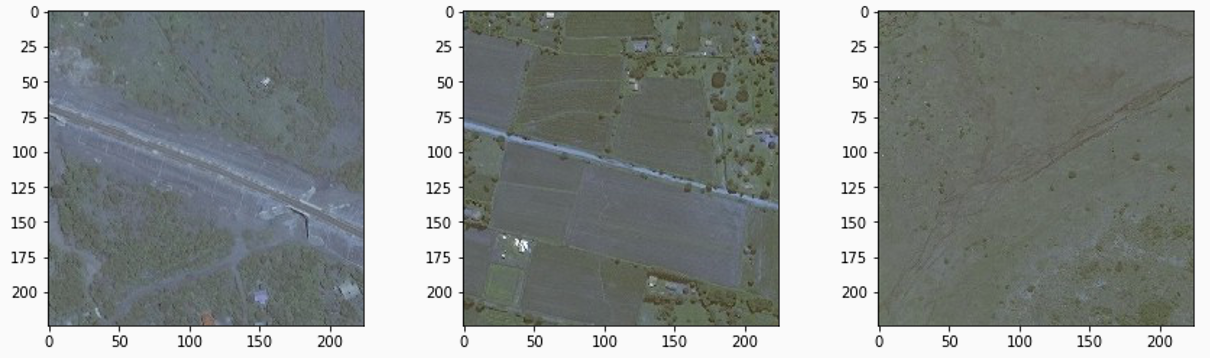}
\caption{Example RGB images of three road quality classes: major, minor, two-track (left to right).}
\label{fig:classImages}
\end{figure}

Our input data consists of 30 cm resolution satellite imagery from DigitalGlobe. Labels consist of road classification and road footprint information from OpenStreetMap. In particular, one road-label pair will contain: one 1000 by 1000 pixel satellite image covering a 300 by 300 squared meter area centered on a given road; a classification of the road belonging to the label set \texttt{\{major, minor, two-track\}}; and geographical connectivity information for the road's constituent GPS coordinates (the polyline). Explicitly, the polyline consists of a set of ordered GPS coordinates which constitute the skeleton of the road shape. OpenStreetMap does \textit{not} provide road polygon information.

\subsection{Collection and Preprocessing}

For Kenya, one point is sampled from among the constituent GPS coordinates of each road. For Peru, two points are sampled in an effort to maintain data parity between the two countries. This sampling procedure results in approximately 170K images with a 1:20:12 \textbf{\texttt{major}:\texttt{minor}:\texttt{two-track}} class ratio for the Kenya distribution, and approximately 97K images with a 13:20:10 \textbf{\texttt{major}:\texttt{minor}:\texttt{two-track}} class ratio for the Peru distribution.

In addition, we seek to augment classification by removing cloudy images from our data. We accomplish this by enforcing an arbitrary threshold on the minimum mean-per-band for each image. We find that a threshold of 150 removes most cloudy images with minimal false positives (i.e. images for which all band means are over 150 are deemed likely to be mostly or completed occluded by clouds, and are therefore removed). We achieve marginal, but noticeable improvement on validation sets with de-clouded data.

In subsequent experiments, we crop and downsize our imagery to 224 x 224 pixels to make training deep convolutional networks minimally viable. In most experiments, we train and evaluate the model on a class-balanced subset of cropped image data to improve convergence and the interpretability of our results. Any divergence from these modeling decision will be indicated where appropriate.


\section{Methods}

In this report we are primarily concerned with maximizing model performance on the multi-class classification task; however, we also aim to establish a more robust set of both quantitative and qualitative considerations to take into account when evaluating the performance of our models. In particular, we perform experiments to answer the following topical questions.

\begin{itemize}
    \item Context dependence: to what extent does the model rely on contextual information to make classifications?
    \item Road segmentation: will a state-of-the-art pretrained road segmentation model be able to extract quality segmentation masks from our data? 
    \item Class separation: are the data uniformly separable or are certain class boundaries more or less distinguished?
    \item Cross-domain transfer: will a model trained on data from one region perform well on data from another?
\end{itemize}

In all our quantitative experiments, we evaluate performance via accuracy (in the case of binary classification) and macro $F_1$ statistics (shown in Equation \ref{f1_macro_eq}). We choose macro-averaging (i.e. class-wise $F_1$) to more equitably represent performance on minority classes in lieu of domination by performance on the majority class.

\begin{equation} \label{precision_eq}
p_i = \frac{TP_i}{TP_i + FP_i}
\end{equation}

\begin{equation} \label{recall_eq}
r_i = \frac{TP_i}{TP_i + FN_i}
\end{equation}

\begin{equation} \label{f1_macro_eq}
F_1 = \frac{1}{C}\sum_{i = 1}^{C}{\frac{p_ir_i}{p_i + r_i}}
\end{equation}

\subsection{Model Training}

As a baseline for deep convolutional classifiers, we choose a standard ResNet50V2 model pretrained \cite{chollet2015keras} on ImageNet \cite{kaiming2016resnetv2}, to which we affix a feed forward prediction network and retrain on our own data. For all models, we optimize using categorical cross-entropy loss and an Adam optimizer with a learning rate of 0.0001. After fine-tuning our baseline ResNetV2 model on de-clouded data, we try other CNN architectures and find that in practice, Xception \cite{DBLP:journals/corr/Chollet16a}, a model based on depthwise separable convolutions, works best. We also find that models are relatively robust, but not impervious to class imbalances. Given that Kenya has a significantly greater class imbalance than Peru, the relatively large difference between unweighted accuracy and $F_1$ in Table $\ref{table:xceptionTable}$ found in Kenya versus Peru is not surprising. However, in both cases, the disparity is relatively minor in comparison to overall model performance.

\begin{table}
\begin{center}
\begin{tabular}{|l|c|c|}
\hline
Country & Unweighted Accuracy & $\text{F}_1$ \\
\hline\hline
Kenya & 0.80 & 0.73 \\
Peru & 0.67 & 0.66 \\
\hline
\end{tabular}
\end{center}
\caption{Performance of Xception model on Kenyan and Peru RGB validation sets.}
\label{table:xceptionTable}
\end{table}

\section{Experiments}

As referenced in our discussion of our methods, our experimental efforts are primarily concerned with characterizing the performance of our model under a variety of constraints and environments.

\subsection{Context Masking}

\begin{table}
\begin{center}
\begin{tabular}{|l|c|c|}
\hline
Masking & Unweighted Accuracy & $\text{F}_1$ \\
\hline\hline
Context-Occluded   & 0.77    &0.72\\
Road-Occluded   & 0.79    &0.71\\
Channel Replacement    & 0.81    &0.74\\
\hline
\end{tabular}
\end{center}
\caption{Performance of Xception model on Kenyan and Peru RGB validation sets.}
\label{table:maskingTable}
\end{table}

\begin{figure}
\centering
\includegraphics[width=\linewidth]{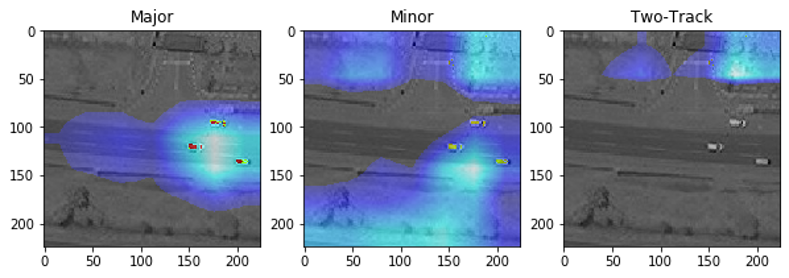}
\caption{Class activation maps generated for a Kenyan major road reveal disparate activation regions depending on the classification.}
\label{fig:camImages}
\end{figure}

We seek to characterize the effect of contextual information on the performance of our road classification models. Specifically, model training on images with relatively small road areas may be dominated by signals from non-road segments of an image. Thus, we seek to generate a binary road mask in order to evaluate the effect of non-road context on road classification.

Our binary mask generation process is as follows. For each Kenyan satellite image, a set of randomly sampled geographical coordinate points within the image are provided (points sampled from OpenStreetMap road footprints). These points are randomly sampled in a larger bounding box (1000 x 1000) from which the image (224 x 224) is later center cropped. Thus, the number of points relevant to the road segment within the cropped image is variable. Following the "thresholding" of these sampled points to within the cropped region, we use the Bresenham line algorithm \cite{Bresenham:1965:ACC:1663347.1663349} to highlight pixels that line segments connecting sequential road points would overlap. From the resulting one-pixel thick road mask, we thicken the mask in order to fully cover the area of the image representing roads. We accomplish this effect by adding pixels within a pre-determined pixel radius of the one-pixel thick road mask. We find that a radius of 20 pixels works well for 1000x1000 images.

We perform a pixel-wise multiplication of our raw satellite image with our generated mask in order to generate context-occluded images. All input images are center-cropped to 224x224 prior to training. In addition to evaluating model trained on models with context-occluded, we also train only-context models with road segments occluded via inverted road masks. Comparison of model performance in differing masking scenarios acts as a proxy for the extent to which our models utilize actual road pixels in classification of the overall image.

Table~\ref{table:maskingTable} summarizes our results from each masking approach. We find that both models achieve similar performance to our non-occluded baseline, with road-occluded models performing slightly better than context-occluded models. These results support the notion that our road classification models are able to classify “roads” with reasonable performance without the actual road being present in the image. Our findings suggest that there is a strong correlation between road type and surrounding geographic context. However, our context-occluded models still achieve reasonable performance, suggesting that the road itself does produce a signal relevant to the model.

As shown in Figure~\ref{fig:camImages}, class activation maps \cite{DBLP:journals/corr/ZhouKLOT15} produced from models trained on non-occluded images further support our findings. While models do seem to focus on relevant road areas of the image for classification of major roads, classification of \texttt{minor} and \texttt{two-track} roads seems to rely more heavily on contextual information.

\subsubsection{Channel Replacement}
A key constraint in the use of pre-trained image classifications models is the allowed number of input channels. Thus, we are motivated to explore the redundancy of RGB channels such that we can substitute alternative, potentially more informative channels to improve model performance. Specifically, in lieu of pixelwise multiplication with binary road masks, we simply replace one RGB input channel with the binary road mask itself. We our motivated by the assumption that most information in our images is preserved in two channels, as well as by the notion that concatenation of binary masks may contribute novel information not present in RGB channels that is useful to our model. We achieve marginal improvement over the standard RGB Xception model, suggesting that our ideal mask input is of greater use to the model than the blue channel. 

While it is certainly possible to try concatenating alternative channels in addition to standard RGB inputs, but we find that in practice, comparable performance to pretrained RGB-only models is difficult to achieve given the large-data requirements of training large, complex models. Thus, in scenarios where there is insufficient data to achieve comparable performance to pretrained models with models trained from scratch, our approach may allow for the inclusion of alternative data sources with greater signal.

\subsubsection{Road Segmentation}

We also assess whether segmentation masks can be used to augment our input data; yet as OSM data do not include segmentation labels and only connectivity information, we cannot train a U-Net of LinkNet architecture on our own data, relying instead on pretrained models. We perform semantic segmentation on a balanced subset of 150 images using the pretrained road extraction model published by the winners of the DeepGlobe 2018 Road Extraction Challenge \cite{Zhou2018DLinkNetLW}. We find that with minimal hyperparameter exploration, the model is not able to produce a minimally viable segmentation for any class of road. On our sampled subset, the model is unable to accurately identify even 50\% of \texttt{major} roads, and even fewer \texttt{minor} and \texttt{two-track} roads. In future experimentation with pretrained segmentation models, we will explore using the raw sigmoid confidences in lieu of forcing a binary mask via some arbitrary threshold. 

\begin{figure}
\centering
\includegraphics[width=0.85\linewidth]{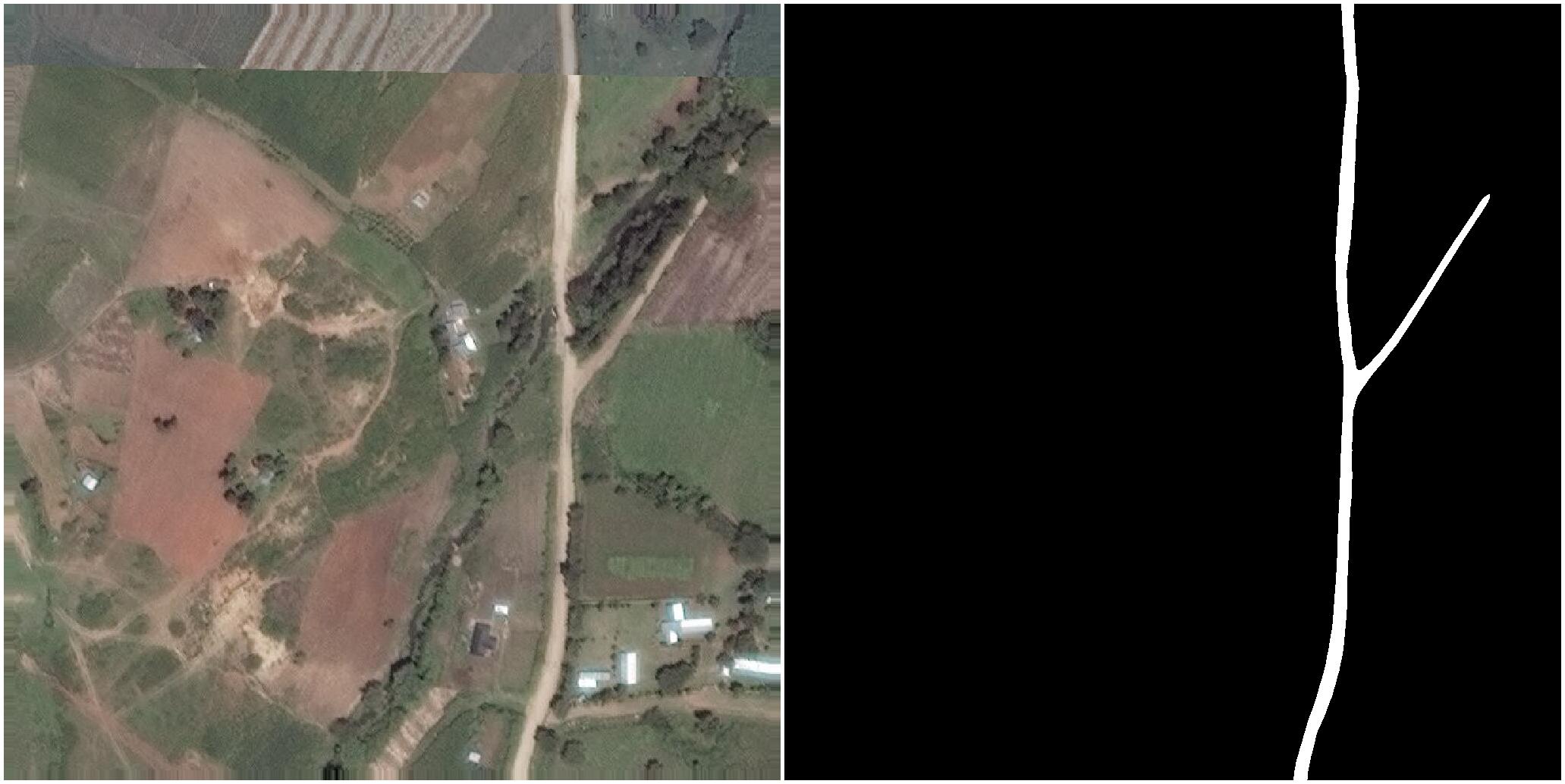}
\caption{A Kenyan two-track road presumed to exist at the center of the image, and the corresponding, incorrect road segmentation mask.}
\label{fig:segmentation}
\end{figure}

This segmentation experiment also challenges the realistic viability of our task and of the assumptions baked into our data. Often, and especially in urban areas, a given satellite image will contain partial segments belonging to more than one road, and in cases where these roads differ in class, the model is forced either to arbitrarily disambiguate between road segments or to become increasingly more dependent on the assumption implicit to our data that the road in question is found directly at the center of the image. This is problematic behavior, as it imposes certain strict constraints on the data our model is able to treat. See Figure \ref{fig:segmentation} for an example of one such adversarial segmentation.

\subsection{Binarization Tests}

We reframe our task as a number of different binary classification problems to assess class-wise separability of our data, and report the results in Table \ref{table:binarization}. Each experiment is conducted with the ResNetV2 architecture on a balanced subset, the labels of which we binarize, with a particular focus on the \texttt{minor} and \texttt{major} classes, which we suspect to be our least and most separable, respectively.

\begin{table}
\begin{center}
\begin{tabular}{|l|c|c|c|}
\hline
Isolated Class & Alternate Class & N & Accuracy \\
\hline\hline
Minor   & Major, Two-track & 15000 & 0.68 \\
Minor   & Two-track & 10000 & 0.70 \\
Minor   & Major & 10000 & 0.88 \\
Major   & Minor, Two-track & 10000 & 0.88 \\
Major   & Two-track & 10000 & 0.92 \\
\hline
\end{tabular}
\end{center}
\caption{Results from binarization experiments.}
\label{table:binarization}
\end{table}

For reference, on a balanced subset of 15K samples with no binarization, a ResNetV2 classifier is only able to achieve a classification accuracy of about 0.70 and an $F_1$ score of \textbf{0.69}. Our suspicion was evidently correct: \texttt{minor} class roads are extremely difficult to distinguish from \texttt{two-track} class roads to the detriment of model performance as a whole, whereas \texttt{major} class roads are relatively easy to distinguish in any situation. However, it cannot be understated how sensitive this result is to the class distribution in the dataset. We train this same binarized classifier (\texttt{major} vs. \texttt{minor}, \texttt{two-track}) on the entire Kenya dataset with class-weighting proportional to the class distribution and find that the model no longer boasts such predictive power, attaining an $F_1$ score of  \textbf{0.63}. This is significantly lower than the result achieved by the 3-class model on balanced data, which, class distribution aside, is the harder problem. This drop in performance is a strong indicator of the need to explore data augmentation methods (e.g. upsampling) to assuage the class imbalance or otherwise revisit data collection methods.

\subsection{Cross-domain Transferability}

\begin{table}
\begin{center}
\begin{tabular}{|l|c|c|}
\hline
Training Domain & Test Domain & Balanced Accuracy \\
\hline\hline
Kenya   & Peru & 0.46\\
Peru   & Kenya & 0.60\\
\hline
\end{tabular}
\end{center}
\caption{Results from cross-domain transfer experiments with Xception.}
\label{table:domainTable}
\end{table}

\begin{figure}
\centering
\includegraphics[width=\linewidth]{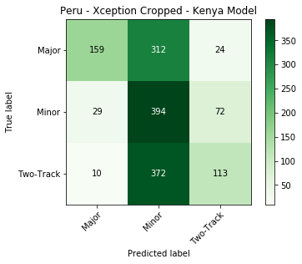}
\includegraphics[width=\linewidth]{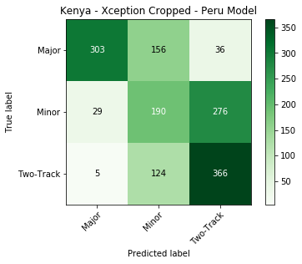}
\caption{Cross-domain confusion matrices on balanced test sets.}
\label{fig:domainMatrix}
\end{figure}

\begin{figure}
\centering
\includegraphics[width=0.45\linewidth]{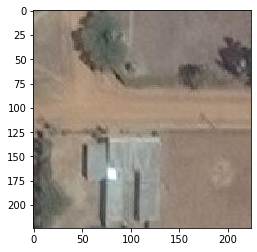}
\includegraphics[width=0.45\linewidth]{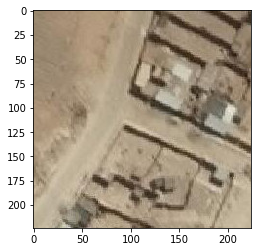}
\caption{A Kenyan minor road classified as a two-track road by Peru-to-Kenya model (left) and a true Peruvian two-track road (right).}
\label{fig:domainImages}
\end{figure}

Finally, we assess the transferability of our models to disparate domains. Given that our models seem to rely heavily on the surrounding geographical context for road classification, we hypothesize that different country domains may drastically affect model performance due to widely varying geographical conditions worldwide. 

Based on Table~\ref{table:domainTable}, we find that there is a significant decrease in model performance when applying Kenya-trained models to Peru, while there is a lesser decrease when applying Peru-trained models to Kenya. Difficulties with model transferability may be characterized via inspection of Figure~\ref{fig:domainMatrix} and Figure~\ref{fig:domainImages}. The variability of performance with direction of transfer also suggests that some countries may have country-specific signals (e.g. Kenya) relevant to classification, while others may have more general signals (e.g. Peru), allowing for training models more ideally-equipped for robust classification across domains. Further, our results suggest that for implementation at scale, road/infrastructure models may require retraining on domains specific to the region of interest, or training on very large, aggregated datasets sampled from all regions of interest.

The performance of our mask-replacement model is comparable, but not superior to the performance of our 3-channel RGB models. This result suggests that there is relatively insignificant data loss when removing entire RGB channels from training.

\section{Conclusion}

We have proposed a method for classification of road quality from satellite imagery with reasonable performance. Our approach is a first step towards both robust classification of roads across the globe, as well as to gaining a more holistic understanding of spatio-temporal road and infrastructure development patterns. We have also revealed several key considerations future satellite imagery based work on infrastructure classification must integrate depending on the purpose of their study. In particular, training under diverse mask occlusion scenarios has highlighted strong correlations between surrounding image context and the road classification. For studies using road classification as a proxy for infrastructure development and other more general economic indicators, the use of surrounding context may not pose a serious issue, and may even be desired. However, in cases where road isolation is required, masking using ground truth labels or trained road segmentation models may be necessary to ensure leakage or dominance of surrounding geographic context into model inference.

The ability of segmentation to aid road classification remains inconclusive due to the inability of pretrained road segmentation models to overcome the slight resolution and image dimension shift to effectively generalize to our domain. However, RGB channel replacement experiments with naively constructed, ground truth road masks suggests high-accuracy segmentation models would only lead to marginal improvement.

Moreover, OpenStreetMap data carries with it some degree of increased variance. The nature of crowd-source, minimally verified and minimally \textit{verifiable} data contributes to the subjectivity and low degree of separability we see in our classification labels for both in-domain and cross-domain comparisons. Our binarization study confirmed that certain classes (\texttt{minor} and \texttt{two-track}), themselves aggregate classes of OSM sub-categorization, are more difficult to distinguish than others. Our cross-domain study proved that classifications are not consistent across regional domains, as evidenced by the variable performance of models evaluated on images from a distribution not present in their training set (Kenya $\rightarrow$ Peru, Peru $\rightarrow$ Kenya).
 
Further, models that allow for the use of surrounding context in road classification may be more susceptible to failure cases in applications to out-of-country or out-of-region applications. Surrounding vegetation, land use, building styles, and other contextual factors are likely to possess at least a weak association with the specific country or region of interest. We have thus highlighted the need for careful curation of datasets used for training road and infrastructure classification models. Specifically, depending on the application, sampling of datasets should occur either within close local proximity of the region of interest, or with diverse multi-region input for more generalized applications. 
 
Future iterations of road classification studies should address several assumptions made by our study. For example, in addition to rotational variation it may be desirable to add translational variation to our data augmentation procedure. This is because while the dataset used for this study ensured the existence of a road intersecting the center of the image, future applications of classification models would likely benefit from models invariant to the fulfillment of this assumption. Continued research will no doubt investigate the viability of OpenStreetMap labels with low resolution image data as we have done here with high resolution image data. Charting the historical expansion of transportation networks necessitates that our algorithm be applicable to satellite image data from the several past decades, which is necessarily lower in resolution. Experimenting with high resolution image data whose resolution we artificially reduce is a logical next step toward this end.




\section{Acknowledgements}

We would like to acknowledge the course faculty and staff for their contributions in the form of persistent, high-quality feedback. In particular we would like to thank Professors Marshall Burke, David Lobell, and Stefano Ermon and course staff Robin Cheong, Matthew Tan, and Burak Uzkent; in addition, we thank Jenny Xue for her assistance with data collection. We would also like to acknowledge Lisa Mandle with the Natural Capital Project and Christina Kennedy with The Nature Conservancy for defining and motivating this work. All map data is copyright OpenStreetMap contributors and available from \url{https://www.openstreetmap.org}.

{\small
\bibliographystyle{ieee}
\bibliography{egbib}
}

\end{document}